\RequirePackage{fix-cm}
\documentclass[twocolumn]{svjour3}          
\smartqed  
\usepackage{graphicx}
\usepackage{mathtools}
\usepackage{hyperref}
\begin{document}

\title{Automatic Exposure Selection and Fusion \\ for High Dynamic Range Photography via Smartphones 
}


\author{Reza Pourreza-Shahri         \and
        Nasser Kehtarnavaz 
}


\institute{R. Pourreza-Shahri \at
              University of Texas at Dallas
           \and
           N. Kehtarnavaz \at
              University of Texas at Dallas
              \email{kehtar@utdallas.edu}           
}

\date{Appeared in Signal, Image and Video Processing volume 11, pages 1437–1444 (2017) \href{https://link.springer.com/article/10.1007/s11760-017-1104-9}{(link)}}

\maketitle

\begin{abstract}
High Dynamic Range (HDR) photography involves fusing a bracket of images taken at different exposure settings in order to compensate for the low dynamic range of digital cameras such as the ones used in smartphones. In this paper, a method for automatically selecting the exposure settings of such images is introduced based on the camera characteristic function. In addition, a new fusion method is introduced based on an optimization formulation and weighted averaging. Both of these methods are implemented on a smartphone platform as an HDR app to demonstrate the practicality of the introduced methods. Comparison results with several existing methods are presented indicating the effectiveness as well as the computational efficiency of the introduced solution.
\keywords{Automatic exposure selection \and High Dynamic Range photography on smartphones \and exposure bracketing}
\end{abstract}

\section{Introduction}
\label{sec:1}
Many camera sensors, in particular the ones used in smartphones, have limited dynamic or contrast ranges. High Dynamic Range (HDR) techniques allow compensating for low dynamic ranges of such sensors by capturing a number of images at different exposures, called an exposure bracket, and by fusing these images to form an HDR image \cite{Ref1}, \cite{Ref2}.\\
Although an exposure bracket may contain many images taken at different exposure settings, Barakat \textit{et al.} \cite{Ref3} showed that in most cases, an exposure bracket consisting of three images is adequate for capturing the full contrast of a scene. Exposure bracketing is thus normally done by taking three images with one taken at the auto-exposure (\textit{AE}) setting together with a brighter looking image and a darker looking image taken at \textit{$EV_{AE} \pm n$} exposure settings, where $n$ is manually selected or is a user-specified exposure level. An automatic exposure selection method makes it possible to generate HDR images without requiring users to select the exposure level \textit{n}. Automatic exposure selection methods in the literature can be grouped into two major categories: scene irradiance-based methods \cite{Ref3}, \cite{Ref4} and ad-hoc methods \cite{Ref5}-\cite{Ref7}.\\
Fusion approaches in the literature can also be placed into one of these two major categories: (i) irradiance mapping \cite{Ref8} followed by tonal mapping, and (ii) direct exposure fusion \cite{Ref9}. Tonal mapping methods work either globally or locally. Global tonal mapping methods, e.g. \cite{Ref10}, use a monotonically increasing curve to compress an irradiance map. Such methods do not retain local image details. On the other hand, local tonal mapping methods retain local image details. Recent tonal mapping algorithms, e.g. \cite{Ref11}-\cite{Ref13}, decompose the luminance of an irradiance map image into a base layer and a detail layer where the base layer consists of large scale variations and the detail layer of small scale variations. Direct exposure fusion methods, e.g. \cite{Ref15}-\cite{Ref21}, use local image characteristics to generate weight maps and then fuse bracket images using such maps. It is worth noting that some exposure fusion works have addressed the generation of HDR images in the presence of changing scenes \cite{Ref22}, \cite{Ref23}.\\
In this paper, a new HDR photography method for scenes that remain static during the time different exposures are captured is introduced and implemented on smartphones. This method consists of two parts: exposure selection and exposure fusion. In the exposure selection part, the scene is analyzed to determine three exposure times automatically for under-exposed, normal-exposed, and over-exposed images. These images are then blended or fused in the exposure fusion part using an optimization framework and weighted averaging.\\
The rest of the paper is organized as follows. In sections 2 and 3, the exposure selection and the exposure fusion parts are described, respectively. Section 4 provides the smartphone implementation or app for generating HDR images. The results and comparisons with several existing methods are then presented in section 5 and finally the paper is concluded in section 6.

\section{Automatic Exposure Selection Method}
\label{sec:2}
\begin{figure}
  \includegraphics[width=0.48\textwidth]{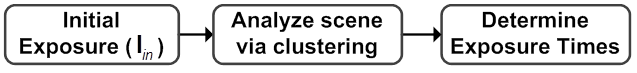}
\caption{Automatic exposure selection steps}
\label{fig:1}
\end{figure}

Exposure bracketing involves the use of three images: a normal-exposed image ($\textbf{I}_{NE}$), an over-exposed image ($\textbf{I}_{OE}$) and an under-exposed image ($\textbf{I}_{UE}$). The exposure selection method we introduced in \cite{Ref7} has been employed here for the smartphone implementation. The steps to find optimal exposure deviations about the normal exposure are shown in Fig.~\ref{fig:1}.

Initially, all the camera parameters are set to automatic and an auto-exposed image $\textbf{I}_{in}$ is captured. The luminance of $\textbf{I}_{in}$ is clustered into dark, normal, and bright regions. These regions represent under-exposed, well-exposed, and over-exposed parts of the image, respectively. The optimal exposure times are then obtained in order to make the dark, the normal, and the bright regions of $\textbf{I}_{in}$ better exposed in $\textbf{I}_{OE}$, $\textbf{I}_{NE}$, and $\textbf{I}_{UE}$, respectively. Unlike the conventional automatic exposure methods that consider a linear relationship between the exposure time and the brightness level, in this work the camera characteristic function is used to establish this relationship and the exposure times are found by mapping the gray-level means of the clustered regions to the optimal gray level (usually 128 for 256-level images). More details of the exposure selection process are provided in \cite{Ref7}.

\section{Fusion Method to Generate HDR Images}
\label{sec:3}
This section describes the fusion of the images $\textbf{I}_{UE}$, $\textbf{I}_{NE}$, and $\textbf{I}_{OE}$, using an algorithm described in the next subsections. Let $\textbf{Y}_{UE}$, $\textbf{Cb}_{UE}$, $\textbf{Cr}_{UE}$, $\textbf{Y}_{NE}$, $\textbf{Cb}_{NE}$, $\textbf{Cr}_{NE}$, $\textbf{Y}_{OE}$, $\textbf{Cb}_{OE}$, $\textbf{Cr}_{OE}$, \textbf{Y}, \textbf{Cb}, \textbf{Cr} represent the Y, Cb, and Cr components of the images $\textbf{I}_{UE}$, $\textbf{I}_{NE}$, $\textbf{I}_{OE}$, and the fused or output image, respectively. Fusion is conducted for the luminance and chrominance component separately as described in the following two subsections.

\subsection{Luminance fusion}
\label{sec:31}

\begin{figure}
  \includegraphics[width=0.48\textwidth]{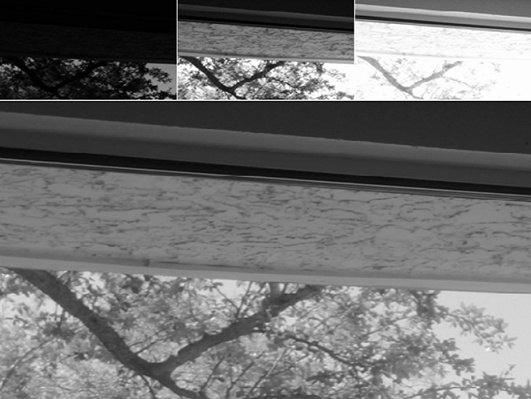}
\caption{Top row from left to right: under-exposed, normal-exposed, and over-exposed images. Bottom row: average image.}
\label{fig:4}
\end{figure}

Here gradient information is used to guide the luminance fusion. The rationale behind using gradient information is that a well-exposed image provides a better representation of edges, that is to say it provides higher gradient values compared to a poorly-exposed image \cite{Ref18}. The gradients are extracted from the well-exposed regions of $\textbf{Y}_{UE}$, $\textbf{Y}_{NE}$, and $\textbf{Y}_{OE}$ and merged into gradient maps along the horizontal ($\mathbf{\Lambda}_h$) and vertical ($\mathbf{\Lambda}_v$) directions. An initial estimate (\textbf{X}) of the fused luminance (\textbf{Y}) is also obtained by averaging $\textbf{Y}_{UE}$, $\textbf{Y}_{NE}$, and $\textbf{Y}_{OE}$. Using the extracted gradients and the initial estimate, the fused luminance (\textbf{Y}) is devised by solving the optimization problem that is stated below.

\begin{equation}
\begin{array}{ll}
\widehat{\textbf{Y}}=\textrm{argmin}_{\textbf{Y}}\big\lbrace
\Vert \textbf{Y}-\textbf{X} \Vert_F^2+
\lambda\Vert \nabla^h\textbf{Y}-\mathbf{\Lambda}_h \Vert_F^2+\\
\qquad \qquad \qquad \enspace \lambda\Vert \nabla^v\textbf{Y}-\mathbf{\Lambda}_v \Vert_F^2
\big\rbrace
\end{array}
\label{eq:10}
\end{equation}

where $\Vert . \Vert_F^2$ denotes the Frobenius matrix norm, $\nabla^h$ and $\nabla^v$ indicate the horizontal and vertical gradient operators, and $\lambda$ is a weighting parameter. The solution to this optimization problem is given by (see Appendix A):

\begin{equation}
\begin{array}{ll}
\widehat{\mathbf{\Psi}}_{u,v}=\\ \quad
\frac{
\mathcal{F}\lbrace\textbf{X}\rbrace_{u,v}+
\lambda\mathcal{F}\lbrace\nabla^h\rbrace_{u,v}^{\dagger}
\mathcal{F}\lbrace\mathbf{\Lambda}_h\rbrace_{u,v}+
\mathcal{F}\lbrace\nabla^v\rbrace_{u,v}^{\dagger}
\mathcal{F}\lbrace\mathbf{\Lambda}_v\rbrace_{u,v}
}
{
1+\lambda\Vert \mathcal{F}\lbrace\nabla^h\rbrace_{u,v} \Vert^2+
\lambda\Vert \mathcal{F}\lbrace\nabla^v\rbrace_{u,v} \Vert^2
}\\
\widehat{\textbf{Y}}=\mathcal{F}^{-1}\big\lbrace\widehat{\mathbf{\Psi}}\big\rbrace
\end{array}
\label{eq:11}
\end{equation}

where \textit{u} and \textit{v} are the pixel coordinates, $\mathcal{F}$ and $\mathcal{F}^{-1}$ denote the forward and inverse Fourier transforms, respectively, and $\dagger$ denotes complex conjugate.\\
For luminance fusion, the images $\textbf{Y}_{UE}$, $\textbf{Y}_{NE}$, and $\textbf{Y}_{OE}$, are available and the output image \textbf{Y} is estimated. As stated earlier, \textbf{X}, $\mathbf{\Lambda}_h$ and $\mathbf{\Lambda}_v$ are obtained from the available image data in the solution provided in (\ref{eq:11}) to estimate \textbf{Y}. In this work, the initial estimate \textbf{X} is chosen to be the average of the images captured at different exposure settings because the average luminance conveys the brightness of a scene as illustrated in Fig.~\ref{fig:4}. In other words, \textbf{X} is set to be

\begin{equation}
\textbf{X}=(\textbf{Y}_{OE}+\textbf{Y}_{NE}+\textbf{Y}_{UE})/3
\label{eq:12}
\end{equation}

The horizontal ($\mathbf{\Lambda}_h$) and the vertical ($\mathbf{\Lambda}_v$) gradients are extracted from the luminance components using the clustering outcome obtained in the exposure selection step. In other words, the gradient in the dark, normal, and bright regions is obtained from the over-exposed, normal-exposed, and under-exposed images, respectively. For this purpose, three binary cluster maps (\textbf{R}) are generated which correspond to the three clusters obtained from the exposure selection step . For a pixel \textit{k} of $\textbf{I}_{in}$, if its gray-level belongs to the cluster 1, 2, or 3 (specified by the dark, normal, and bright areas, respectively), then the corresponding pixel in $\textbf{R}_{UE}$, $\textbf{R}_{UE}$, or $\textbf{R}_{UE}$ is set to 1, where $\textbf{R}_{UE}$, $\textbf{R}_{NE}$, and $\textbf{R}_{OE}$ represent the binary cluster maps for the under-exposed, normal-exposed, and over-exposed clusters, respectively. Hence, for a pixel \textit{k}

\begin{equation}
\begin{split}
\textbf{R}_{UE}^k=\left\{\begin{array}{ll}
1,\quad C_{\textbf{I}_{in}^k}=1\\
0,\quad \textrm{otherwise}
\end{array} \right.\\
\textbf{R}_{NE}^k=\left\{\begin{array}{ll}
1,\quad C_{\textbf{I}_{in}^k}=2\\
0,\quad \textrm{otherwise}
\end{array} \right.\\
\textbf{R}_{OE}^k=\left\{\begin{array}{ll}
1,\quad C_{\textbf{I}_{in}^k}=3\\
0,\quad \textrm{otherwise}
\end{array} \right.
\end{split}
\label{eq:13}
\end{equation}

where $C_{\textbf{I}_{in}^k}$ denotes the cluster label of the $k^{th}$ pixel of $\textbf{I}_{in}$. Then, $\mathbf{\Lambda}_h$ and $\mathbf{\Lambda}_v$ for the pixel \textit{k} are computed as follows:

\begin{equation}
\begin{array}{ll}
\mathbf{\Lambda}_h^k=
\textbf{R}_{UE}^k\big(\nabla^h\textbf{Y}_{OE}\big)^k+
\textbf{R}_{NE}^k\big(\nabla^h\textbf{Y}_{NE}\big)^k+\\
\qquad\enspace\textbf{R}_{OE}^k\big(\nabla^h\textbf{Y}_{UE}\big)^k\\
\mathbf{\Lambda}_v^k=
\textbf{R}_{UE}^k\big(\nabla^v\textbf{Y}_{OE}\big)^k+
\textbf{R}_{NE}^k+\big(\nabla^v\textbf{Y}_{NE}\big)^k+\\
\qquad\enspace\textbf{R}_{OE}^k\big(\nabla^v\textbf{Y}_{UE}\big)^k
\end{array}
\label{eq:14}
\end{equation}

The method described above causes sharp transitions on cluster borders since image gradients get multiplied by binary cluster maps. In order to avoid such sharp transitions, a low-pass filter is applied to every binary map. Here, the so-called guided filter \cite{Ref13} is applied to the binary masks to make transitions of cluster borders smooth. In other words, the smooth cluster maps ($\mathbf{\Omega}$) for the three clusters are obtained as follows:

\begin{equation}
\begin{array}{ll}
\mathbf{\Omega}_{UE}=\textrm{guided\_filter}(\textbf{X},\textbf{R}_{UE},r,\varepsilon)\\
\mathbf{\Omega}_{NE}=\textrm{guided\_filter}(\textbf{X},\textbf{R}_{NE},r,\varepsilon)\\
\mathbf{\Omega}_{OE}=\textrm{guided\_filter}(\textbf{X},\textbf{R}_{OE},r,\varepsilon)
\end{array}
\label{eq:15}
\end{equation}

where \textbf{X} acts as a guide and \textit{r} and $\varepsilon$ indicate the filter parameters. The smoothed cluster maps are then normalized. The normalized weights for a pixel \textit{k} are computed this way

\begin{equation}
\begin{array}{ll}
\omega_{UE}^k=\mathbf{\Omega}_{UE}^k\big/\big( \mathbf{\Omega}_{UE}^k+\mathbf{\Omega}_{NE}^k+\mathbf{\Omega}_{OE}^k\big)  \\
\omega_{NE}^k=\mathbf{\Omega}_{NE}^k\big/\big( \mathbf{\Omega}_{UE}^k+\mathbf{\Omega}_{NE}^k+\mathbf{\Omega}_{OE}^k\big)  \\
\omega_{OE}^k=\mathbf{\Omega}_{OE}^k\big/\big( \mathbf{\Omega}_{UE}^k+\mathbf{\Omega}_{NE}^k+\mathbf{\Omega}_{OE}^k\big)
\end{array}
\label{eq:16}
\end{equation}

Once the normalized weights are computed, $\mathbf{\Lambda}_h$ and $\mathbf{\Lambda}_v$ are obtained as follows:

\begin{equation}
\begin{array}{ll}
\mathbf{\Lambda}_h^k=
\omega_{UE}^k\big( \nabla^h\textbf{Y}_{OE}\big)^k+
\omega_{NE}^k\big( \nabla^h\textbf{Y}_{NE}\big)^k+\\
\qquad\enspace\omega_{OE}^k\big( \nabla^h\textbf{Y}_{UE}\big)^k \\
\mathbf{\Lambda}_v^k=
\omega_{UE}^k\big( \nabla^v\textbf{Y}_{OE}\big)^k+
\omega_{NE}^k\big( \nabla^v\textbf{Y}_{NE}\big)^k+\\
\qquad\enspace\omega_{OE}^k\big( \nabla^v\textbf{Y}_{UE}\big)^k
\end{array}
\label{eq:17}
\end{equation}

\textbf{X}, $\mathbf{\Lambda}_h$ and $\mathbf{\Lambda}_v$ computed via (\ref{eq:12}) and (\ref{eq:17}) are then inserted into (\ref{eq:11}) to obtain \textbf{Y}.

\subsection{Chrominance fusion}
\label{sec:32}

Unlike the luminance component, chrominance fusion is achieved through weighted averaging. The saturation value of a pixel is used as a weight to fuse its chrominance components. The saturation $s$ for a pixel \textit{k} with red, green, and blue components of $r_k$, $g_k$, $b_k$ is computed this way

\begin{equation}
\begin{array}{ll}
\rho_k=(r_k+g_k+b_k)/3 \\
s_k=\sqrt{(r_k-\rho_k)^2+(g_k-\rho_k)^2+(b_k-\rho_k)^2}
\end{array}
\label{eq:18}
\end{equation}

Let $s_{UE}^k$, $s_{NE}^k$, and $s_{OE}^k$ represent the saturation of the pixel \textit{k} in $\textbf{I}_{UE}$, $\textbf{I}_{NE}$, and $\textbf{I}_{OE}$, respectively. The normalized weights are obtained as follows:

\begin{equation}
\begin{array}{ll}
\varpi_{UE}^k=s_{UE}^k\big/\big( s_{UE}^k+s_{NE}^k+s_{OE}^k \big)\\
\varpi_{NE}^k=s_{NE}^k\big/\big( s_{UE}^k+s_{NE}^k+s_{OE}^k \big)\\
\varpi_{OE}^k=s_{OE}^k\big/\big( s_{UE}^k+s_{NE}^k+s_{OE}^k \big)
\end{array}
\label{eq:19}
\end{equation}

And the chrominance components of the pixel \textit{k} are then fused according to these equations:

\begin{equation}
\begin{array}{ll}
\textbf{Cb}^k=\varpi_{UE}^k\textbf{Cb}_{UE}^k+\varpi_{NE}^k\textbf{Cb}_{NE}^k+\varpi_{OE}^k\textbf{Cb}_{OE}^k\\
\textbf{Cr}^k=\varpi_{UE}^k\textbf{Cr}_{UE}^k+\varpi_{NE}^k\textbf{Cr}_{NE}^k+\varpi_{OE}^k\textbf{Cr}_{OE}^k
\end{array}
\label{eq:20}
\end{equation}

\section{Smartphone Implementation}
\label{sec:4}
A smartphone app is developed in this work to demonstrate the practicality aspect of the introduced automatic exposure selection and fusion methods. This app is developed for smartphones running Android operating system by utilizing the following software tools: Android Studio and Android Native Development Kit (Android NDK) which allows incorporating C/C++ codes into Android Studio. The entire processing is divided into two parts: capturing part, which is done via Java code, and processing part, which is done via C code. More details about the two parts are provided in the following two subsections.

\subsection{Capturing}
\label{sec:41}
The capturing part involves taking $\textbf{I}_{in}$ followed by taking $\textbf{I}_{OE}$, $\textbf{I}_{NE}$, and $\textbf{I}_{UE}$. A button named Capture is provided in the app to launch capturing images. Once the \textit{Capture} button is pressed by the user, the capturing process starts as described in Algorithm \ref{tab:2}. The output of Algorithm \ref{tab:2} consists of three images in the YCbCr format, corresponding to the under-exposed, normal-exposed, and over-exposed conditions as well as the clustering outcome in the form of three binary maps.

\begin{table}
\caption{Capturing part}
\label{tab:2}
\begin{tabular}{l}
\hline\noalign{\smallskip}
\textbf{Input}: None.\\
\textbf{Output}: Three YCbCr images and binary maps\\
\enspace 1. Perform an initial capture\\
\enspace 2. Cluster the luminance component of this initial \\
\enspace \enspace \enspace  capture, record the 3 cluster means and generate \\ \enspace \enspace \enspace  three binary maps\\
\enspace 3. Initialize three \textit{CaptureRequests} and set the\\ \enspace \enspace \enspace \textit{CaptureRequest} exposure times\\
\enspace 4. Initialize a burst \textit{CaptureSession} with the three \\ \enspace \enspace \enspace defined \textit{CaptureRequests}\\
\enspace 5. Launch \textit{CaptureSession} and wait for camera to \\ \enspace \enspace \enspace acquire the three images\\
\enspace 6. Pass the three images together with the three binary 
\\ \enspace \enspace \enspace maps through JNI (Java Native Interface) to the C \\ \enspace \enspace \enspace  fusion code\\
\noalign{\smallskip}\hline
\end{tabular}
\end{table}

\subsection{Fusion part}
\label{sec:42}

When the three images are made available from the capturing part, the processing outlined in section~\ref{sec:3}, which is coded in C, is performed and an HDR image as described in Algorithm \ref{tab:3} is generated. This image is then sent back to the Java environment for getting saved in memory.

\begin{table}
\caption{Fusion part}
\label{tab:3}
\begin{tabular}{l}
\hline\noalign{\smallskip}
\textbf{Input}: Three images in YCbCr format and binary maps.\\
\textbf{Output}: HDR image\\
\enspace 1. Fuse luminance component\\
\enspace \enspace a. Obtain \textbf{X} via (\ref{eq:12})\\
\enspace \enspace b. Calculate the horizontal and vertical gradients of \\ \enspace \enspace \enspace  \enspace $\textbf{Y}_{UE}$, $\textbf{Y}_{NE}$, and $\textbf{Y}_{OE}$\\
\enspace \enspace c. Apply the guided filter to get smooth cluster maps  \\ \enspace \enspace \enspace \enspace via (\ref{eq:15}) and normalize the weights via (\ref{eq:16})\\
\enspace \enspace d. Acquire the gradients via (\ref{eq:17})\\
\enspace \enspace e. Estimate \textbf{Y} via (\ref{eq:11})\\
\enspace 2. Fuse chrominance component\\
\enspace \enspace a. Obtain the saturation for each pixel of $\textbf{I}_{OE}$, $\textbf{I}_{NE}$,\\ \enspace \enspace \enspace \enspace and $\textbf{I}_{UE}$ via (\ref{eq:18}) and calculate the normalized \\ \enspace \enspace \enspace \enspace chrominance weights via (\ref{eq:19})\\
\enspace \enspace b. Calculate \textbf{Cb}, \textbf{Cr} via (\ref{eq:20})\\
\enspace 3. Combine \textbf{Y}, \textbf{Cb}, and \textbf{Cr} to form the output image \\ \enspace \enspace \enspace in the YCbCr format and then convert to the RGB \\ \enspace \enspace \enspace format\\
\noalign{\smallskip}\hline
\end{tabular}
\end{table}

\section{Results and Discussion}
\label{sec:5}

This section provides the results of the experimentations conducted to evaluate the introduced automatic exposure selection and fusion methods for generating HDR images.\\
An Android app  was designed to capture 34 images of a scene at 8M size and fixed ISO, fixed lens position, with the auto-white-balance enabled, across different exposure times. 34 exposure times varying from 0.5 sec to 1/4000 sec in steps of 1/3 standard exposure value were used to generate a ground-truth irradiance map image ($\textbf{HDR}_{ref}$) of the scene using the method introduced in \cite{Ref8}. The above process was repeated for 10 different indoor and outdoor scenes. The complete dataset is made available for public use at http://www.utdallas.edu/$\sim$kehtar/ImageEnhancement/\\HDR/Scenes.\\

The first set of experiments mentioned next addressed selecting the optimal parameters while the second and third sets of experiments addressed the performance and the results obtained by a modern smartphone, respectively.

\subsection{Parameter selection}
\label{sec:51}

\begin{figure}
  \includegraphics[width=0.48\textwidth]{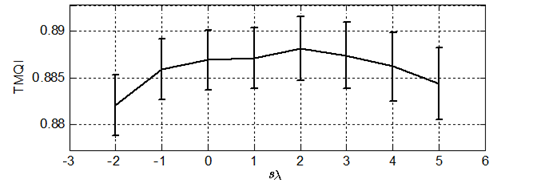}
\caption{TMQI vs. scale $s_{\lambda}$}
\label{fig:7}
\end{figure}

The parameters of the exposure fusion part included the regularization parameter $\lambda$ of the optimization in (\ref{eq:11}), as well as the guided filter parameters, i.e. the filter radius \textit{r} and the regularization parameter $\varepsilon$ in (\ref{eq:15}). For the guided filter, since maximal smoothing within short distances was desired, the minimal value of 1 for \textit{r} together with a large value of 1 for $\varepsilon$ were considered. The kernels $[-1, 1]$ and $[-1, 1]^T$ were used as the horizontal and vertical gradient operators, respectively.\\
The experiments revealed that the optimal $\lambda$ was highly dependent on the image size. To address this issue, resized copies of the images were generated by scaling the dimension from 0.1 to 1 in steps of 0.1. The fusion algorithm was then applied to all the images, the original images and resized ones. $\lambda$ was varied according to (\ref{eq:21}) and the widely used image quality measure of Tone-Mapped Quality Index TMQI \cite{Ref24} was computed for every $\lambda$ ,

\begin{equation}
\lambda=SIZE/2^{s_{\lambda}},s_{\lambda}=-2,-1,\ldots,5
\label{eq:21}
\end{equation}

where $SIZE$ denotes the diameter of the input image and $s_{\lambda}$ is a scaling factor. The average TMQI over all the images for different scale values is shown in Fig.~\ref{fig:7}. This figure indicates that the scale value of 2 generated the best TMQI. Hence, $\lambda$ was set to $\lambda = SIZE/4$ for the subsequent experimentations.

\begin{figure}
  \includegraphics[width=0.48\textwidth]{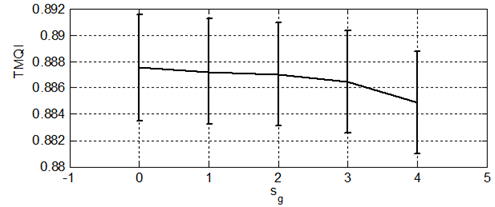}
\caption{TMQI vs. scale $s_g$}
\label{fig:8}
\end{figure}

Furthermore, in order to gain computational efficiency, another experiment was conducted by applying the guided filtering operation to the down-sampled binary maps (instead of the original size) and TMQI was recomputed. The binary cluster maps were scaled to a smaller size, after applying the guided filter, the filtering outcome was rescaled to the original size. All the other steps of the exposure fusion were kept the same. The dimensions of the binary cluster maps were lowered as follows:

\begin{equation}
SCALE=1/2^{s_g},s_g=0,1,\ldots,4
\label{eq:22}
\end{equation}

where $SCALE$ denotes the down-sampling ratio and $s_g$ represents a scaling factor. The average TMQI over all the images for different $s_g$ values is shown in Fig.~\ref{fig:8}. As can be seen from this figure, TMQI dropped monotonically as $s_g$ was increased. However, since the decrease in TMQI was more as $s_g$ varied from 3 to 4 compared to the previous decreases, $s_g=3$ was chosen in the smartphone app. In other words, for computational efficiency reasons on a smartphone platform, the binary clustered maps were down-sampled with a factor of 8 before applying the guided filter.

\subsection{Comparison}
\label{sec:52}

\begin{figure*}
  \includegraphics[width=0.99\textwidth]{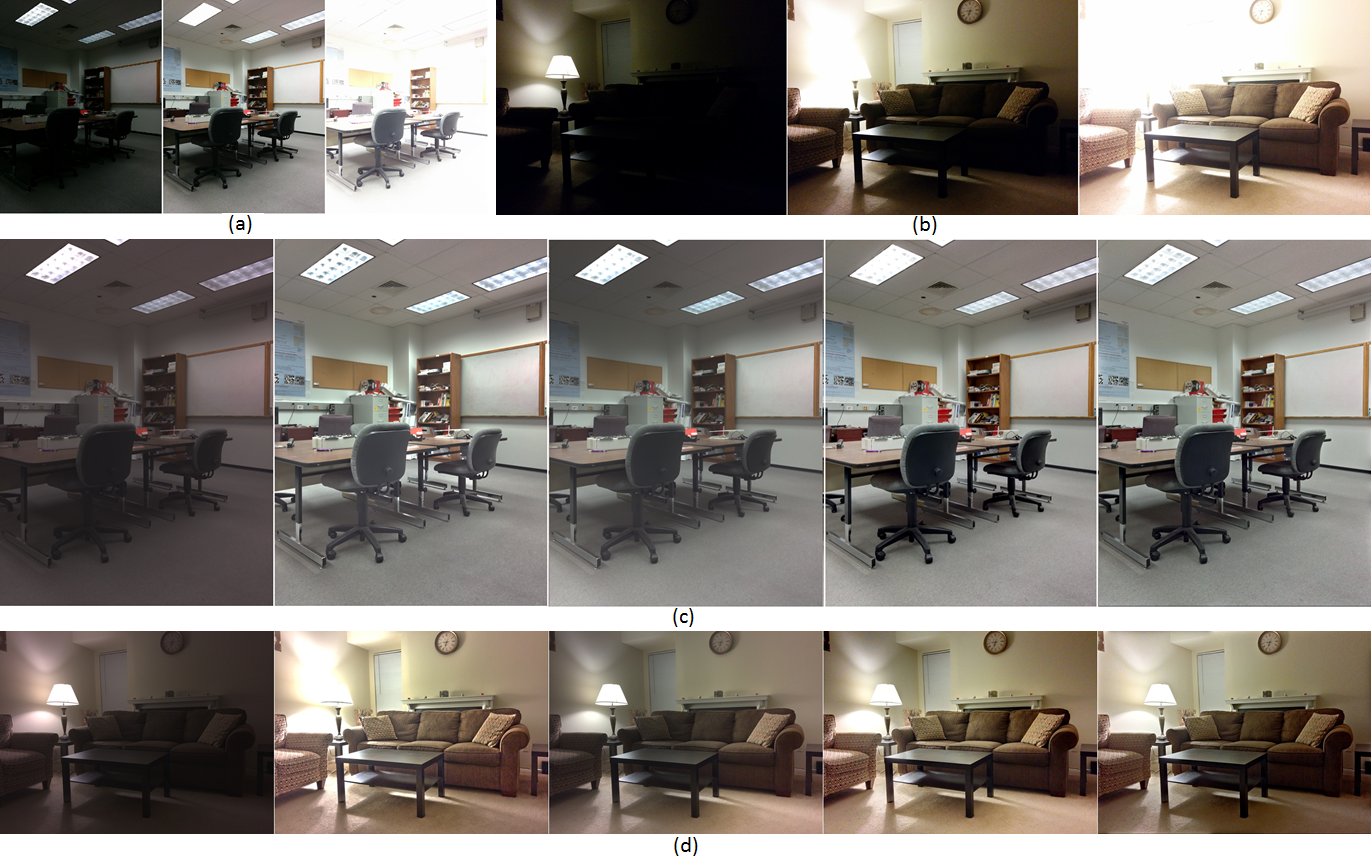}
\caption{Sample results, (a) sequence 1, (b) sequence 2, from left to right: under-exposed, normal-exposed, and over exposed, (c) HDR image  for sequence 1, (d) HDR image for sequence 2, from left to right: iCam, WLS, GF, EF, OB.}
\label{fig:9}
\end{figure*}

This section provides objective and subjective comparisons with several representative existing methods. The method introduced in this paper, named OB, together with iCam \cite{Ref10}, Weighted Least Squares (WLS) \cite{Ref11}, Guided Filtering (GF) \cite{Ref13}, and Exposure Fusion (EF) \cite{Ref15} methods were applied to the image sequences. Sample image sequences as well as the corresponding fusion results are displayed in Fig.~\ref{fig:9}. As can be seen from Figs.~\ref{fig:9} (c) and (d), the fusion results using the EF and OB methods were found to be better compared to the other methods. However, the light region in Fig.~\ref{fig:9} (c) appeared over-exposed and the wall region above the light in Fig.~\ref{fig:9} (d) lost some color information when using the EF method whereas these effects did not occur when using the OB method.

\begin{figure}
  \includegraphics[width=0.48\textwidth]{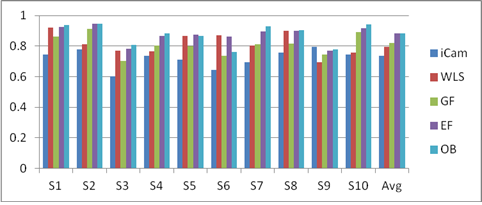}
\caption{TMQI for the scenes together with the average TMQI}
\label{fig:10}
\end{figure}

The TMQI measure averaged over all the 10 scenes are provided in Fig.~\ref{fig:10}. As can be seen from this figure, the performance of the EF and OB methods were found comparable and higher than those of the other methods.

\begin{figure}
  \includegraphics[width=0.48\textwidth]{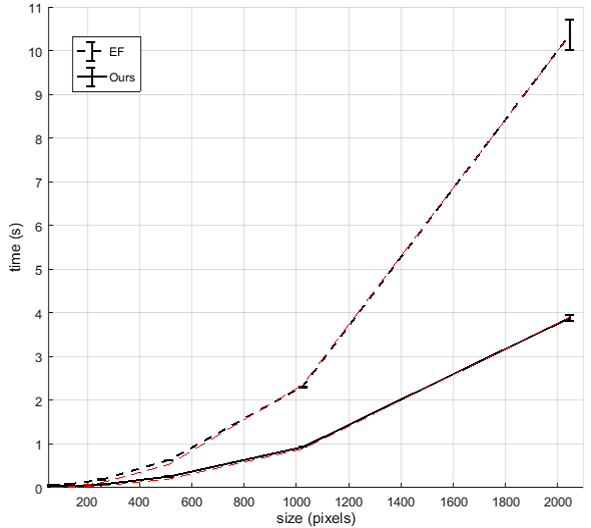}
\caption{Running time comparison}
\label{fig:11}
\end{figure}

The processing times of the aforementioned methods were measured by applying them to the images in the dataset resized to 1 megapixels. These times were found to be 2.67, 22.61, 1.07, 2.37, and 0.99 seconds for the iCam, WLS, GF, EF, and OB methods, respectively, when using a 2.66 GHz machine with all the methods coded in Matlab. It is also worth mentioning that the processing times for the iCam, WLS, and GF methods, which all fall under the tonal mapping category of HDR, only reflect the time associated with the tonal mapping and do not include the time associated with  the scene irradiance map generation. Since the TMQI measure for the iCam, WLS, and GF methods was found to be lower compared to those for the EF and OB methods, the processing time comparison was limited to the EF and OB methods which is stated below.\\
The computational complexities of the EF and OB methods are $C_1O\big(N^2log(N)\big)$ and $C_2O\big(N^2log(N)\big)$, respectively, where $C_1$ and $C_2$ denote constants and $N$ denotes the image size (height or width). Although the two methods have the same big $O$, the constants $C_1$ and $C_2$ have a noticeable impact on the processing time as $N$ increases. Fig.~\ref{fig:11} shows the average processing times of the EF and OB methods over the resized versions of the captured scenes in terms of the image size expressed in pixels. The processing times were measured for $N$=64, 128, 256, 512, 1024, 2048. In addition, $C_1N^2log(N)$ and $C_2N^2log(N)$ curves were fitted to the processing times thus generating the values of 8.47e-8 and 2.25e-7 for the constants $C_1$ and $C_2$, respectively. As can be seen from this figure, the two dashed lines fit the processing times closely. As a result, the processing time of the OB method is on average $C_2/C_1=0.37$ that of the EF method, that is the OB method is about 60\% faster than the EF method. More computational gain can be achieved noting that the Fourier transforms of $\nabla^h$ and $\nabla^v$ in (\ref{eq:11}) are input image-independent and can be computed offline and stored in memory.

\begin{figure}
  \includegraphics[width=0.48\textwidth]{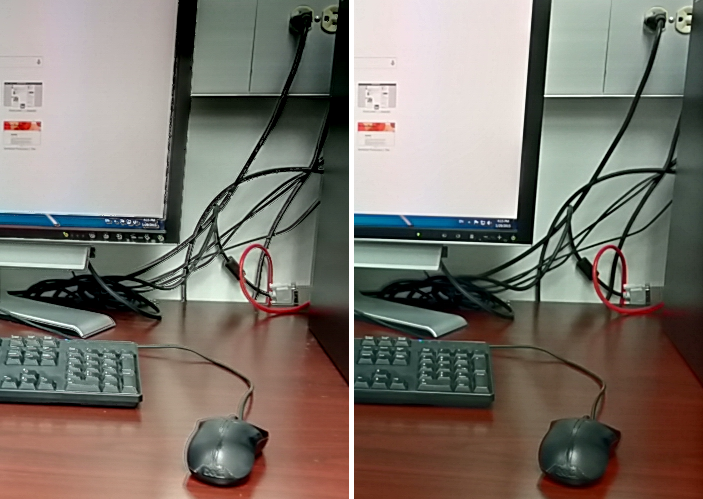}
\caption{Misalignment effect; left: EF method, right our OB method.}
\label{fig:12}
\end{figure}

Although the introduced solution is intended for use in scenes that do not change during the time that different exposure images are captured, the exposure selection and capturing parts are designed and implemented in such a way that there is minimal delay in capturing the three exposure images. Thus, in practice, one experiences little or no misalignments between the three exposure images because of possible changes that may occur in a scene. Furthermore, the exposure fusion part is tolerant to slight misalignments since it extracts the fusion information locally from the auto-exposed image rather than the other exposure images. An example of an HDR image in the presence of misalignments that are generated by the EF and OB methods is shown in Fig.~\ref{fig:12}. As can be seen from this figure, the ghost effect is noticeable around the cables and the edges of the monitor and the mouse in the image generated by the EF method while no ghost artifacts are present in the image generated by the OB method. It is worth noting that it is possible to use computationally efficient image registration techniques, e.g. \cite{Ref4}, \cite{Ref25}, to align different exposure images before performing fusion for situations where severe misalignments may occur.

\subsection{Smartphone results}
\label{sec:53}

\begin{figure}
  \includegraphics[width=0.48\textwidth]{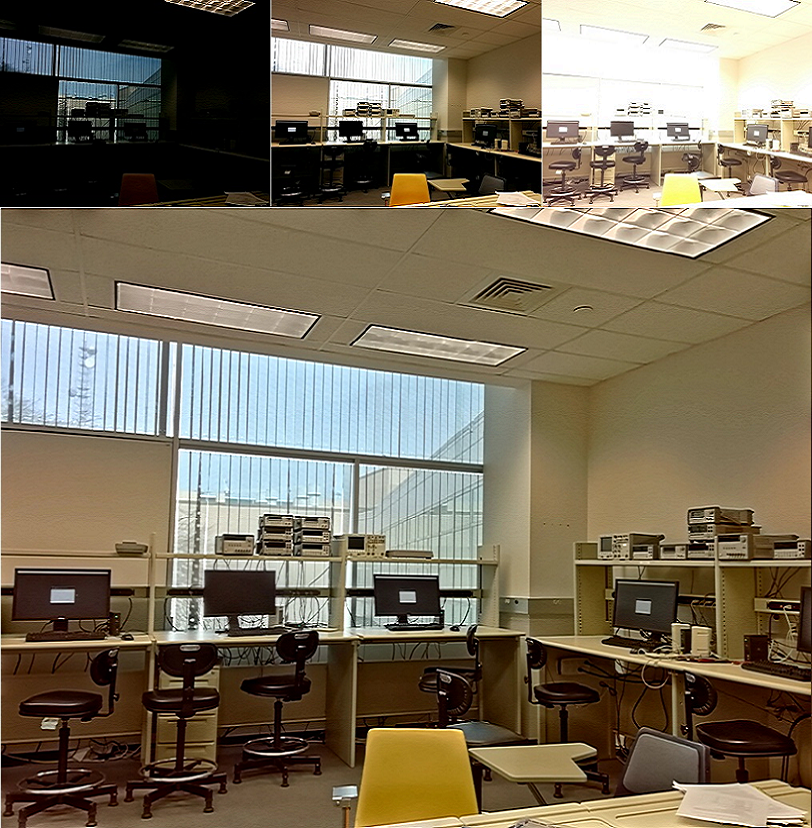}
\caption{Under-exposed, normal-exposed, and over-exposed images together with their fusion outcome.}
\label{fig:14}
\end{figure}

In this subsection, the actual smartphone implementation outcome of the OB method is reported. This HDR app can be downloaded from http://www.utdallas.edu/\\$\sim$kehtar/HDRApp.apk  and run on an Android smartphone. A sample exposure selection and fusion outcome obtained by running the developed app on a smartphone is shown in Fig.~\ref{fig:14} together with the corresponding under-exposed, normal-exposed, and over-exposed images.

The average processing time for images of size $768\times1024$ pixels on a modern smartphone was found to be about 1.8s. Note that the intention here has been to show the practicality of running the introduced solution on smartphones. It should be realized that it is possible to achieve lower processing times as the app code running on the smartphone does not utilize vector processing on the NEON coprocessor present on nearly all modern smartphones. 

\section{Conclusion}
\label{sec:6}

A method for automatic exposure selection and a method for fusion of exposure bracket images were introduced in this paper. The exposure selection was done by analyzing the brightness of a scene via clustering and the camera characteristic function. For exposure fusion, the luminance and chrominance of three bracket images were blended via an optimization formulation and weighted averaging, respectively. The smartphone implementation and comparisons with the existing methods have shown the practicality and performance of the introduced solution.

\appendix
\section{Optimization Solution}

This appendix provides the solution of the optimization problem with only one gradient term. The derivation for two terms is straightforward and not included here to save space.  The optimization formulation for one gradient term is given by 

\begin{equation}
\widehat{\textbf{Y}}=\textrm{argmin}_{\textbf{Y}}\big\lbrace
\Vert \textbf{Y}-\textbf{X} \Vert_F^2+
\lambda\Vert \nabla\textbf{Y}-\mathbf{\Lambda} \Vert_F^2\big\rbrace
\label{eq:23}
\end{equation}

where $\mathbf{\Lambda}$ represents the gradient of \textbf{Y} and $\nabla$ indicates the gradient operator. In vector form, Equation (\ref{eq:23}) can be written as:

\begin{equation}
\widehat{\textbf{y}}=\textrm{argmin}_{\textbf{y}}\big\lbrace
\Vert \textbf{y}-\textbf{x} \Vert^2+
\lambda\Vert \textbf{C}\textbf{y}-\mathbf{\delta} \Vert^2\big\rbrace
\label{eq:24}
\end{equation}

where \textbf{y}, \textbf{x}, and $\mathbf{\delta}$ represent the column vector versions of \textbf{Y}, \textbf{X}, and $\mathbf{\Lambda}$, respectively, and \textbf{C} denotes the block-circulant matrix representation of $\nabla$. By taking the derivative with respect to \textbf{y}, the following solution is obtained

\begin{equation}
\widehat{\textbf{y}}=
\big(\textbf{I}+\lambda\textbf{C}^T\textbf{C} \big)^{-1}
\big(\textbf{x}+\lambda\textbf{C}^T\mathbf{\delta} \big)
\label{eq:25}
\end{equation}

where \textbf{I} denotes the identity matrix. Since \textbf{C} is a block-circulant matrix, it can be represented in diagonal form as:

\begin{equation}
\textbf{C}=\textbf{W}\textbf{E}\textbf{W}^{-1}
\label{eq:26}
\end{equation}

where \textbf{E} is the diagonal version of \textbf{C} and \textbf{W} is the DFT (Discrete Fourier Transform) operator. The diagonal values of \textbf{E} correspond to the DFT coefficients of $\nabla$ ($\nabla$  should be zero-padded properly before applying DFT). Hence, (\ref{eq:25}) can be rewritten as:

\begin{equation}
\widehat{\textbf{y}}=
\big(\textbf{I}+\lambda\textbf{W}\textbf{E}^{\dagger}\textbf{E}\textbf{W}^{-1} \big)^{-1}
\big(\textbf{x}+\lambda\textbf{W}\textbf{E}^{\dagger}\textbf{W}^{-1}  \mathbf{\delta} \big)
\label{eq:27}
\end{equation}

By multiplying both sides of (\ref{eq:27}) by $\textbf{W}^{-1}$, the following equation is resulted

\begin{equation}
\textbf{W}^{-1}\widehat{\textbf{y}}=
\big(\textbf{I}+\lambda\textbf{E}^{\dagger}\textbf{E} \big)^{-1}
\big(\textbf{W}^{-1}\textbf{x}+\lambda\textbf{E}^{\dagger}\textbf{W}^{-1}  \mathbf{\delta} \big)
\label{eq:28}
\end{equation}

Since $\textbf{W}^{-1}\widehat{\textbf{y}}$, $\textbf{W}^{-1}\textbf{x}$, and $\textbf{W}^{-1}\mathbf{\delta}$  correspond to the DFT of $\widehat{\textbf{y}}$, \textbf{x}, and $\mathbf{\delta}$, respectively, (\ref{eq:28}) can be stated as follows:

\begin{equation}
\mathcal{F}\lbrace\widehat{\textbf{Y}}_{u,v}\rbrace=\frac{
\mathcal{F}\lbrace\textbf{X}\rbrace_{u,v}+
\lambda\mathcal{F}\lbrace\nabla\rbrace_{u,v}^{\dagger}
\mathcal{F}\lbrace\mathbf{\Lambda}\rbrace_{u,v}
}
{
1+\lambda\Vert \mathcal{F}\lbrace\nabla\rbrace_{u,v} \Vert^2
}
\label{eq:29}
\end{equation}



\begin{thebibliography}{}
\bibitem{Ref1}
Mann, S., Picard, R.,: Being `undigital' with digital cameras: extending dynamic range by combining differently exposed pictures. In: IS\&T, 48th annual conference, USA, pp. 422-428 (1995).
\bibitem{Ref2}
Reinhard, E., Heidrich, W., Debevec, P., Pattanaik, S., Ward, G., Myszkowski, K.: High Dynamic Range Imaging, 2nd edn. Morgan Kaufmann Publishers, San Francisco (2010).
\bibitem{Ref3}
Barakat, N., Hone, A.N., Darcie, T.E.: Minimal-bracketing sets for high-dynamic-range image capture. IEEE Trans. Image Process. \textbf{17}(10), 1864-1875 (2008).
\bibitem{Ref4}
Gupa, M., Iso, D., Nayar, S.K.: Fibonacci exposure bracketing for high dynamic range imaging. In: IEEE Intl. Conf. Cumputer Vision (ICCV), Australia, pp. 1473-1480 (2013).
\bibitem{Ref5}
Huang, K., Chiang, J.: Intelligent exposure determination for high quality HDR image generation. In: IEEE Intl. Conf. Image Process. (ICIP), Australia, pp. 3201-3205 (2013).
\bibitem{Ref6}
Pourreza-Shahri, R., Kehtarnavaz, N.: Automatic exposure selection for high dynamic range photography. In: IEEE Intl. Conf. Consumer Electronics (ICCE), USA, pp. 469-497, (2015).
\bibitem{Ref7}
Pourreza-Shahri, R., Kehtarnavaz, N.: Exposure bracketing via automatic exposure selection. In: IEEE Intl. Conf. Image Proc. (ICIP), Canada, pp. 320-323 (2015).
\bibitem{Ref8}
Debevec, P., Malik, J.: Recovering high dynamic range radiance maps from photographs. In: 24th Annual Conf. Computer Graphics and Interactive Techniques (SIGGRAPH), USA, pp. 369-378 (1997).
\bibitem{Ref9}
Cvetković, S., Klijn, J., With, P.H.N.: Tone-mapping functions and multiple-exposure techniques for high dynamic-range images. IEEE Trans. Consumer Electron. \textbf{54}(2), 904-911 (2008).
\bibitem{Ref10}
Kuang, J., Johnson, G.M., Fairchild, M.D.: iCAM06: A refined image appearance model for HDR image rendering. J. Visual Communication \textbf{18}, 406-414 (2007).
\bibitem{Ref11}
Farbman, Z., Fattal, R., Lischinski, D., Szeliski, R.:Edge-preserving decompositions for multi-scale tone and detail manipulation. ACM Trans. Graphics \textbf{21}(3), 249–256 (2008).
\bibitem{Ref12}
Durand, F., Dorsey, J.: Fast bilateral filtering for the display of high-dynamic-range images. ACM Trans. Graphics \textbf{21}(3), 257–266 (2002).
\bibitem{Ref13}
He, K., Sun, J., Tang, X.: Guided image filtering. IEEE Trans. Pattern Analysis and Machine Intelligence \textbf{35}(6), 1397-1409 (2013).
\bibitem{Ref15}
Mertens, T., Kautz, J., Van Reeth, F.:Exposure fusion: a simple and practical alternative to high dynamic range photography. Computer Graphics Forum \textbf{28}(1), 161–171 (2009).
\bibitem{Ref16}
Li, S., Kang, X.: Fast multi-exposure image fusion with median filter and recursive filter. IEEE Trans. Consumer Electron. \textbf{58}(2), 626–632 (2012).
\bibitem{Ref17}
Song, M., Tao, D., Chen, C., Bu, J., Luo, J., Zhang, C.: Probabilistic exposure fusion. IEEE Trans. Image Process. \textbf{21}(1), 341– 357 (2012).
\bibitem{Ref18}
Zhang, W., Cham, W.-K.: Gradient-directed multi-exposure composition. IEEE Trans. Image Process. \textbf{21}(4), 2318–2323 (2012).
\bibitem{Ref19}
Shen, R., Cheng, I., Shi, J., Basu, A.: Generalized random walks for fusion of multi-exposure images. IEEE Trans. Image Process. \textbf{20}(12), 3634-3646 (2011).
\bibitem{Ref20}
Shen, R., Cheng, I., Basu, A.: QoE-based multi-exposure fusion in hierarchical multivariate gaussian CRF. IEEE Trans. Image Process. \textbf{22}(6), 2469–2478 (2013).
\bibitem{Ref21}
Xu, L., Du, J., Zhang, Z.: Feature-based multiexposure image-sequence fusion with guided filter and image alignment. J. Electron. Imaging \textbf{24}(1), 013022 (2015).
\bibitem{Ref22}
Hu, J., Gallo, O., Pulli, K.: Exposure stacks of live scenes with hand-held cameras. In: Europ. Conf. Computer Vision (ECCV), Italy, pp. 499-512 (2012).
\bibitem{Ref23}
Tico, M., Gelfand, N., Pulli, K.: Motion-blur-free exposure fusion. In: IEEE Intl. Conf. Image Process. (ICIP), China, pp. 3321-3324 (2010).
\bibitem{Ref24}
Yeganeh H., Wang, Z.: Objective quality assessment of tone-mapped images. IEEE Trans. Image Process. \textbf{22}(2), 657-667 (2013).
\bibitem{Ref25}
Ward, G.: Fast, robust image registration for compositing high dynamic range photographs from handheld exposures. J. Graphic Tools \textbf{8}(2), 17-30 (2003).

\end{thebibliography}
\end{document}